\documentclass[final,3p,times,twocolumn]{elsarticle}
\usepackage{etoolbox}
\makeatletter
\patchcmd{\ps@pprintTitle}{\footnotesize\itshape
       Preprint submitted to \ifx\@journal\@empty Elsevier
       \else\@journal\fi\hfill\today}{\relax}{}{}
\makeatother
\pdfoutput=1
\usepackage{blindtext, graphicx, amsmath, algorithm, algpseudocode, pifont, algcompatible, comment, layout, amsthm, amssymb}
\usepackage{enumitem}   

\usepackage{float}
\usepackage[utf8]{inputenc}
\usepackage[english]{babel}

\journal{ICT Express}

\begin{document}

\begin{frontmatter}

\title{DeepImageSpam: Deep Learning based Image Spam Detection  }
%\author{Sayantan Chowdhury\corref{cor1}} 
%\address{Department of Electronics \& %Telecommunication, Jadavpur University, India}
%\cortext[cor1]{Corresponding author:}
%\ead{sc.juetce@gmail.com}
\author{Amara Dinesh Kumar\corref{cor1}}
\ead{dineshkumar.amara@gmail.com}
\address{Department of Electronics and Communication Engineering, Amrita School of Engineering, Coimbatore,\\ Amrita Vishwa Vidyapeetham, India}
\author{Vinayakumar R\corref{cor1}}
\author{Soman KP}
\address{Center for Computational Engineering and Networking (CEN), Amrita School of Engineering, Coimbatore,\\ Amrita Vishwa Vidyapeetham, India}

\begin{abstract}
 Hackers and spammers are employing innovative and novel techniques to deceive novice and even knowledgeable internet users. Image spam is one of such technique where the spammer varies and changes some portion of the image such that it is indistinguishable from the original image fooling the users. This paper proposes a deep learning based approach for image spam detection using the convolutional neural networks which uses a dataset with 810 natural images and 928 spam images for classification achieving an accuracy of 91.7\% outperforming the existing image processing and machine learning techniques.

\end{abstract}
%% with the increase in the usage of internet the 
\begin{keyword}
%% keywords here, in the form: keyword \sep keyword
Image Spam \sep Deep Learning \sep Spam Detection \sep CNN 
%% MSC codes here, in the form: \MSC code \sep code
%% or \MSC[2008] code \sep code (2000 is the default)
\end{keyword}

\end{frontmatter}

%%
%% Start line numbering here if you want
%%
% \linenumbers

%% main text
\section{Introduction}
\label{sec1}
The Internet has become a second self for most of the people today and many of our financial transactions, social interactions and communication are dependent on it and not always completely safe. Intruders, hackers and attackers are always in the quest for exploiting the users by hacking, spamming and impersonating. One of the major common and cost-effective and targeted attack in recent years is sending spam to users. spam detection and classification are very active fields of research and attackers are finding new ways to spam the users even with multiple layers of security mechanisms. They are finding the bugs and vulnerabilities and exploiting them actively improving on day to day basis. It is very common that users receive emails impersonating banks and other authorities asking for other personal details. Spam messages are causing huge loss to organizations. Many resources are getting wasted like mail server space, network bandwidth, spam filtering, mail server processing.

Attackers may redirect users to counterfeit products sites, impersonate an authenticated site and stealing sensitive information, spread fake news and providing wrong information, false marketing. spam messages are wasting the time and effort of the users and distracting them thus decreasing the productivity flooding email and forcing to delete and clean it frequently.

Image spam is altering the Image spam is increasing in the recent years with tremendous growth. A major reason for that is many of the email clients are filtering the spam text emails the subject sender and email content.but when it comes to the image detecting the spam is not easy particularly when attackers are embedding text into the images.
Generally, email spam is detected using the spam filters which are evolved state now and can detect most of the spam with high accuracy but when it comes to image spam detection it is still in nascent stage and active research is going on for detecting with high accuracy.
Initially Image spam in the form of HTML and to counter it researchers started using the OCR techniques. Attackers then used captcha based techniques obfuscating the text in the images but still readable by the humans but difficult to identify for an algorithm. This problem motivated researchers to use image processing techniques for image spam identification.

\subsection{what are spam filters}
Spam filters either block the spam messages or send them to a spam folder. The spam filter analyses the sender, subject, metadata and other related information of the mail and them classify it as spam or legitimate.
They can be individual spam filters maintaining a block list containing addresses where you can add or delete from the respective list.
In a community-based spam filtering, all can collaborate and add entries to the block list for improving the performance and ease of use. Initially set of conditions are programmed as rules for spam filters and later machine learning is being employed in latest spam filter which is proved more efficient and accurate.

\section{Related Work}
\label{sec2}
\subsection{Image processing techniques}
\subsubsection{optical character recognition}
In OCR the text is extracted from the image using the image processing techniques like edge detection and then obtained text is passed to a conventional text spam filter which detects and classify the spam image\cite{6496446}.

spammers have applied various image processing techniques like changing foreground and background also changing text font,size and colour made the OCR based method obsolete\cite{biggio2011survey}.

\subsubsection{Colour Histograms}

The colour histograms of normal images are mostly continuous and spam images contain isolated peaks in the histogram using which we can detect the spam image.But the accuracy is low and it is not a reliable method\cite{BIGGIO20111436}.
Histogram of gradients(HOG) is one more approach used commonly which works similar to colour histograms
using the edge detection of the image and then plotting intensity plot can identify the spam image\cite{6375110}.

\subsection{Using machine learning techniques}
For performing classification using the machine learning algorithms the features are to be selected and extracted manually.Their are two types of features in the image spam detection and they are high level features and low level features.

steps for the attackers for sending the image spam are first developing the template of the image and then obfuscating it randomly and send to to different users which made identification of the image spam difficult and reason for lower detection accuracy.Features are further classified into  Low level features are High level features\cite{Mehta:2008:DIS:1367497.1367565}.
Features like  sender,meta data,message header are extracted and training dataset is prepared and labeled.
\subsubsection{Support Vector Machines (SVM)}
Support vector machines(SVM's) are the most used machine learning algorithms for the image spam detection and because of high accuracy and robustness to misclassifications is the reason researcher prefer SVM\cite{krasser2007identifying}.
SVM is a supervised learning algorithm used for the classification of data.It consists of support vectors which divide and classifies the data.It classifies the non linear data using kernel trick in which the non linear data is projected to higher dimensions to make it linearly separable by a plane which is generally referred as hyper plane.It does this using a kernel function and their are different types of kernel functions like linear,polynomial,radial basis function(RBF) and Gaussian kernels.
Image spam detection is a binary classification problem and two classes are spam and not spam.Using the training data that is collected and labeled according to respective classes model is trained and then the model is tested by giving the test data and performance of model is evaluated.

The major drawback of this method is we have to manual extract the high level and low level features and feed to the classifier which is a time consuming task.

other machine learning classifiers like Logistic regression,Naive Bayes are also used for the image spam detection but SVM outperforms them by comparatively giving better performance. 

\subsection{using Deep learning techniques}
Initially, neural networks are used for image spam detection and then now research has shifted focus on applying the deep learning algorithms.
Deep learning consists of neural network layers which automatically extracts the features from the data in hierarchical pattern and then predicts and classifies the data.
\section{Convolutional neural networks}
Convolutional neural networks (CNN's) are one of the highly efficient deep learning algorithms used for classifying data (particularly image data) using supervised learning technique.
They consist of an Input layer and convolution layer followed by pooling layer and again convolution and pooling layers alternatively based on the size and architecture of the network. The final layer is a fully connected layer. Fully connected layer converts the final scalar outputs of individual classes into their respective probabilities using a non-linear activation function and commonly softmax is used at the last laye\cite{8126009}.
CNN's are commonly used for the image processing applications as it can process the spatial information effectively capturing the pixel-related information using the convolution on to the image with strides\cite{Vinayakumar2018}.

\section{Data set} 
The dataset used in the experiment consist of  928 spam images and 810 normal images which collected from different sources\cite{gao2008image} and all are RGB images in various dimensions which in preprocessing are reshaped to 56$\times$56 images.The sample images are shown in figure.1 and figure.2

Dataset is subdivided into both training and testing datasets. Training dataset consists of 742 spam images and 648 normal images.Testing dataset consists of 186 spam images and 162 normal images.

\begin{figure}
\includegraphics[width=7.2cm, height=6cm]{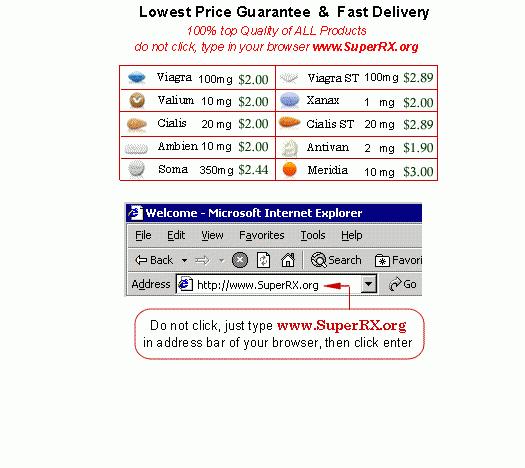}
\caption{Sample spam image from dataset}
\end{figure}
\begin{figure}
\includegraphics[width=7.2cm, height=6cm]{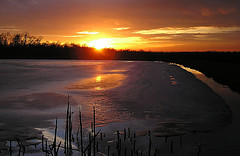}
\caption{Sample non-spam image from dataset}
\end{figure}

\section{Experiments} 
%CNN
The images are normalized and then given to the model for training.First convolution layer the kernel size used is 3$\times$3 with input shape 32$\times$56$\times$56 with RELU (Rectified Linear Unit) activation function in the first convolution layer and then with max pooling layer of size  32$\times$27$\times$27 we are down sampling the data to half of the original dimension and subsequent layers follow the similar pattern.The brief description of the CNN layers architecture along with the output shape is described in table 1.drop out is 0.25 which means we randoms abandon some of the weights to avoid the over fitting and it acts as regularization.Batch size of 32 is used for training each epoch and model is trained for total 1000 epochs.

\begin{table}[h]
\centering
\begin{tabular}{|l|l|} 
\hline
\textbf{Layer (type)~~}         & \textbf{Output Shape~}  \\ 
\hline
conv2d\_1 (Conv2D)~             & (None, 32, 56, 56)      \\ 
\hline
activation\_1 (Activation)      & (None, 32, 56, 56)      \\ 
\hline
conv2d\_2 (Conv2D)              & (None, 32, 54, 54)      \\ 
\hline
activation\_2 (Activation)      & (None, 32, 54, 54)~     \\ 
\hline
max\_pooling2d\_1 (MaxPooling2) & (None, 32, 27, 27)      \\ 
\hline
dropout\_1 (Dropout)~~          & (None, 32, 27, 27)      \\ 
\hline
conv2d\_3 (Conv2D)              & (None, 64, 27, 27)      \\ 
\hline
activation\_3 (Activation)      & (None, 64, 27, 27)~     \\ 
\hline
conv2d\_4 (Conv2D)              & (None, 64, 25, 25)      \\ 
\hline
activation\_4 (Activation)~     & (None, 64, 25, 25)      \\ 
\hline
max\_pooling2d\_2 (MaxPooling2  & (None, 64, 12, 12)      \\ 
\hline
dropout\_2 (Dropout)            & (None, 64, 12, 12)      \\ 
\hline
flatten\_1 (Flatten)            & (None, 9216)            \\ 
\hline
dense\_1 (Dense)                & (None, 128)             \\ 
\hline
activation\_5 (Activation)      & (None, 128)~            \\ 
\hline
dropout\_3 (Dropout)~           & (None, 128)             \\ 
\hline
dense\_2 (Dense)~               & (None, 1)               \\ 
\hline
activation\_6 (Activation)      & (None, 1)               \\
\hline
\end{tabular}
\caption{CNN architecture description}
\end{table}

\section{Results}
\label{sec3}
Total images are split in ratio of 80 percent training data and 20 percent testing data.The model is evaluated after the convolutional neural network is trained on training dataset.It is then tested on the testing data set and result metrics accuracy,precision,recall and f1score are mentioned in the table 2.binary entropy loss and adam optimizer were used for training and checkpoints were saved periodically in the hdf file.
\begin{table}[h]
\centering
\begin{tabular}{|c|c|}
\hline
\textbf{metrics} & \textbf{percentage} \\ \hline
accuracy         & 0.917               \\ \hline
recall           & 0.857               \\ \hline
precision        & 1.000               \\ \hline
f1score          & 0.923               \\ \hline
\end{tabular}
\caption{Results metrics evaluated on test data set} \label{results}
\end{table}
The training and testing are performed on a distributed computing cluster platform with i7 cpu processor and 8 GB RAM system configuration.Keras,Sklearn and Tensorflow deeplearning libraries are used for training and testing.

%\vspace{-0.3cm}
\section{Conclusion}
\label{sec4}

In this research we have used the convolutional neural network(CNN) which is a deep learning network architecture for image spam detection.The deep learning approach gives better accuracy when compared with the machine learning and other conventional image processing based methods and also avoids the manual feature extraction task by automatically identifying the features by itself reducing the time and effort.Binary classification of image is performed the model is trained with existing labelled data set and then tested with the test data then metrics are evaluated.Further research can be carried out by exploring other deep learning algorithms like RNN and LSTM and tuning the architecture and hyper parameters may provide interesting insights.Capsule networks can also be tested on the data set which are giving promising results recently when compared with the convolutional neural networks for image related techniques\cite{kumar2018novel}.

%% References
%%
%% Following citation commands can be used in the body text:
%% Usage of \cite is as follows:
%%   \cite{key}         ==>>  [#]
%%   \cite[chap. 2]{key} ==>> [#, chap. 2]
%%

%% References with bibTeX database:

\bibliographystyle{elsarticle-num}

\vspace{-0.3cm}

\bibliographystyle{unsrt}
\bibliography{ref}
\end{document}